\documentclass{article}

\usepackage[final]{neurips_2020}

\usepackage[utf8]{inputenc} % allow utf-8 input
\usepackage[T1]{fontenc}    % use 8-bit T1 fonts
\usepackage[hidelinks]{hyperref}
\usepackage{url}            % simple URL typesetting
\usepackage{booktabs}       % professional-quality tables
\usepackage{amsfonts}       % blackboard math symbols
\usepackage{nicefrac}       % compact symbols for 1/2, etc.
\usepackage{microtype}      % microtypography

\title{Data and its (dis)contents: A survey of dataset development and use in machine learning research}
\author{
Amandalynne Paullada\\
Department of Linguistics\\
University of Washington 

\And Inioluwa Deborah Raji\\
Mozilla Foundation

\And Emily M. Bender\\
Department of Linguistics\\
University of Washington

\And Emily Denton\\
Google Research 

\And Alex Hanna\\
Google Research }

\begin{document}

\maketitle

\begin{abstract}
    Datasets have played a foundational role in the advancement of machine learning research. They form the basis for the models we design and deploy, as well as our primary medium for benchmarking and evaluation. Furthermore, the ways in which we collect, construct and share these datasets inform the kinds of problems the field pursues and the methods explored in algorithm development. However, recent work from a breadth of perspectives has revealed the limitations of predominant practices in dataset collection and use. In this paper, we survey the many concerns raised about the way we collect and use data in machine learning and advocate that a more cautious and thorough understanding of data is necessary to address several of the practical and ethical issues of the field.
\end{abstract}

\section{Introduction}

The importance of datasets for machine learning research cannot be overstated. Datasets have been seen as the limiting factor for algorithmic development and scientific progress \citep{halevy2009unreasonable, sun2017revisiting}, and a select few benchmark datasets have shaped some of the most significant developments in the field. Benchmark datasets have also played a critical role in orienting the goals, values, and research agendas of the machine learning community \citep{Dotan2020}.

In recent years, machine learning systems have been reported to achieve `super-human' performance when evaluated on benchmark datasets, such as the GLUE benchmark for English textual understanding \citep{wang2019glue}. However, recent work that has surfaced the shortcomings of such datasets as meaningful tests of human-like reasoning ability reveals how this appearance of progress may rest on faulty foundations.

As the machine learning field increasingly turned to data-driven approaches, the sort of skilled and methodical human annotation applied in dataset collection practices in earlier eras was spurned as `slow and expensive to acquire', and a turn toward unfettered collection of increasingly large amounts of data from the Web, alongside increased reliance on non-expert crowdworkers, was seen as a boon to machine learning \citep{halevy2009unreasonable, imagenet_cvpr09}. These data practices tend to abstract away the human labor, subjective judgments and biases, and contingent contexts involved in dataset production. However, these details are important for assessing whether and how a dataset might be useful for a particular application, for enabling better, more systematic error analysis, and for acknowledging the significant difficulty required in constructing useful datasets.  Enormous scale has been mythologized as beneficial to generality and objectivity, but all datasets have limitations and biases \citep{boyd2012critical}. 

The data-driven turn in AI research \,---\,which has placed large scale datasets at the center of model development and evaluation\,---\, makes a careful, critical review of the datasets and practices of dataset creation and use crucial for at least two reasons: first, as systems trained in this way are deployed in real-world contexts that affect the lives and livelihoods of real people, it is essential that researchers, advocacy groups and the public at large understand both the contents of the datasets and how they affect system performance. Second, as the culture of the field has focused on benchmarks as the primary tool for both measuring and driving research progress \citep{Schlangen2020TargetingTB}, understanding what they are measuring (and how well) becomes increasingly urgent.

We conduct a survey of the literature of recent issues pertaining to data in machine learning research, with our primary focus on work in computer vision and natural language processing. We structure our survey around three themes. The first (\S\ref{sec:design}) deals with studies which critically review the design of the datasets used as benchmarks. This includes studies which audit existing datasets for bias (\S\ref{sec:contents}), those which examine existing datasets for spurious correlations which make the benchmarks gameable (\S\ref{sec:spurious}), those which critically analyze the framing of tasks (\S\ref{sec:task}), and work promoting better data collection and documentation practices (\S\ref{sec:doc}). Next, we review approaches to improving these aspects of datasets, while still maintaining the fundamental research paradigm (\S\ref{sec:use}). In looking at approaches to filtering and augmenting data and modeling techniques aimed at mitigating the impact of bias in datasets, we see further critiques of the current state of datasets. However, we find that these approaches do not fully  address the broader issues with data use. Finally, we survey work on dataset practices as a whole, including critiques of their use as performance targets (\S\ref{sec:bm}), perspectives on data management (\S\ref{sec:dm}) and reuse (\S\ref{sec:du}), and papers raising legal issues pertaining to data collection and distribution (\S\ref{sec:legal}).

\section{Definitions}

We follow \cite{Schlangen2020TargetingTB} in distinguishing between \emph{benchmarks}, \emph{tasks}, \emph{capabilities}, and \emph{datasets}. While his work focused on natural langauge processing, we broaden these defintions to include aspects of other machine learning applications. In this context, a \emph{task} is constituted of an input space and output space and an expected mapping between them. Schlangen notes that there are typically both \emph{intensional} and \emph{extensional} definitions of tasks. An intensional definition describes the relationship between input and output (e.g.\ the output in automatic speech recognition is a transcription of the audio signal in the input), where an extensional definition is simply the set of input-output pairs in the dataset. Thus tasks are exemplified by \emph{datasets}, i.e.\ sets of input-output pairs that conform, if valid, to the intensional definition of the task. Tasks can be of interest for two (not mutually exclusive) reasons: Either they map directly into a use case (e.g.\ automatic transcription of audio data) or they illustrate cognitive \emph{capabilities}, typical of humans, that we are attempting to program into machines. In the former case, a task is suitable as a \emph{benchmark} (for comparing competing systems to each other) if the task is well-aligned with its real-world use case and the dataset is sufficiently representative of the data the systems would encounter in production. In the latter case, establishing the value of the task as a benchmark is more involved: as Schlangen argues, success on the task has to be shown to rely on having some set of capabilities that are definable outside of the task itself and transferable to other tasks.

\section{Dataset design and development} 
\label{sec:design}
\begin{quote}
``Every data set involving people implies subjects and objects, those who collect and those who make up the collected. It is imperative to remember that on both sides we have human beings.''

-- Mimi \d{O}n\d{u}\d{o}ha, `The Point of Collection'\footnote{\citet{onuoha2016point}}
\end{quote}

In this section, we review papers which explore issues with the contents of datasets that arise due to the manner in which they were collected, the assumptions guiding the dataset construction process and the set of questions guiding their development.

\subsection{Representational concerns}
\label{sec:contents}
In recent years there has been growing concern regarding the degree and manner of representation of different sociodemographic groups within prominent machine learning datasets. For example, a glaring under-representation of darker skinned subjects has been identified within prominent facial analysis datasets \citep{gendershades} and the images in object recognition datasets have been overwhelmingly sourced from Western countries \citep{DeVries19}. \citet{zhao-etal-2018-gender} found a stark underrepresentation of female pronouns in the commonly used OntoNotes dataset for English coreference resolution. Stereotype-aligned correlations have also been identified in both computer vision and natural language processing datasets. For example, word co-occurrences in natural language processing datasets frequently reflect social biases and stereotypes relating to race, gender, (dis)ability, and more \citep{garg2018word, Hutchinson2020} and correlations between gender and activities depicted in computer vision datasets have been shown to reflect common gender stereotypes \citep{zhao-etal-2017-men, Burns2018WomenAS, van2016stereotyping}. \citet{dixon2018measuring} found that a dataset for toxicity classification contained a disproportionate association between words describing queer identities and text labeled as `toxic.' In an examination of the person categories within the ImageNet dataset \citep{imagenet_cvpr09}, \cite{excavatingai} uncovered millions of images of people that had been labelled with offensive categories, including racial slurs and derogatory phrases. In a similar vein, \cite{Prabhu2020LargeID} examined a broader swath of image classification datasets that were constructed using the same categorical schema as ImageNet, finding a range of harmful and problematic representations, including non-consensual and pornographic imagery of women. In response to the work of \cite{excavatingai}, a large portion of the ImageNet dataset has been removed \citep{Yang2020}. Similarly, \citeauthor{Prabhu2020LargeID}'s [\citeyear{Prabhu2020LargeID}] examination prompted the complete removal of the TinyImages dataset \citep{tinyimages}.

\subsection{Exposing spurious cues exploited by ML models}
\label{sec:spurious}

While deep learning models have seemed to achieve remarkable performance on challenging tasks in artificial intelligence, recent work has illustrated how these performance gains may be due largely to `cheap tricks'\footnote{To borrow a term from \citet{levesque2014our}} rather than human-like reasoning capabilities. \citet{geirhos2020shortcut} illustrate how deep neural networks rely on \textit{shortcuts}, or decision rules that do not extrapolate well to out-of-distribution data and are often based on incidental associations, for performance. Oftentimes, these shortcuts arise due to annotation artifacts in datasets that allow models to overfit to training data and to rely on nonsensical heuristics to `solve' the task.  Recent work has revealed the presence of shortcuts in commonly used datasets that had been conceived of as proving grounds for particular competencies, such as `reading comprehension' and other `language understanding' capabilities. Experiments that illuminate data artifacts, or `dataset ablations' as \citet{heinzerling2019cleverhans} calls them, involve  simple or nonsensical baselines such as training models on incomplete inputs and comparing performance to models trained on full inputs. Much recent work in NLP has revealed how these simple baselines are competitive, and that models trained on incomplete inputs for argument reasoning, natural language inference, and reading comprehension \,---\, i.e., tasks structured such that no human could do much more than randomly guess the correct output \,---\, perform quite well \citep{niven-kao-2019-probing, gururangan-etal-2018-annotation, poliak-etal-2018-hypothesis, kaushik-lipton-2018-much}\footnote{\citet{storks2019recent} and \citet{schlegel2020beyond} provide more comprehensive reviews of datasets and dataset ablations for natural language inference.}. \citet{sugawara2020assessing} show that models trained on instances from machine reading comprehension datasets with scrambled words can still get the `right' answer. Many of these issues result from the assumptions made in task design and in the underspecification of instructions given to human data labelers, and can thus can be addressed by rethinking the format that dataset collection takes. In light of this, recent work has proposed critical approaches to designing annotation frameworks to leverage human `common sense' \citep{srivastava2020robustness} and more critical approaches to reading comprehension dataset creation and use \citep{gardner-etal-2019-making} to pre-empt spurious correlations. 

\subsection{How do datasets legitimize certain problems or goals?}
\label{sec:task}

As the previous sections have laid out, the mapping between inputs and `gold' labels contained in datasets is not always a meaningful one, and the ways in which tasks are structured can lead models to rely on faulty heuristics for making predictions. The problems this raises aren't limited to misleading conclusions based on benchmarking studies: When machine learning models can leverage spurious cues to make predictions well enough to beat a baseline in the test data, the resulting systems can appear to legitimize spurious tasks that do not map to real world capabilities. \citet{jacobsen2020shortcuts} point out that shortcuts in deep learning, as described in Section \ref{sec:spurious}, make ethically dubious questions seem answerable, and advise, `When assessing whether a task is solvable, we first need to ask: should it be solved? And if so, should it be solved by AI?' Simply because a mapping can be learned does not mean it is meaningful, and as we review in Section \ref{sec:spurious}, this mapping can rely on spurious correlations in the data. 

Decisions about what to collect in the first place and the `problematization' that guides data collection leads to the creation of datasets that formulate pseudoscientific, often unjust tasks. For example, several papers in recent years that attempt to predict attributes such as sexuality and other fluid, subjective personal traits from photos of human faces presuppose that these predictions are possible and worthwhile to make. But these datasets, like those discussed above, enable a reliance on meaningless shortcuts. These in turn support the apparent `learnability' of the personal traits in question: an audit by \citet{aguera2018algorithms} found that a model trained on the `gaydar' data was really learning to spot stereotypical choices in grooming and self-expression, which are by no means universal, while \cite{Gelman_2018} discuss how such a study strips away context and implies the existence of an ``essential homosexual nature.'' The task rests on a pseudoscientific essentialism of human traits. Another example, from NLP, is GermEval 2020 Task 1 \citep{GermEval2020}, which asked systems to reproduce a ranking of students by IQ scores and grades using only German short answer texts produced by the students as input. By setting up this task as feasible (for machine models or otherwise), the task organizers suggested that short answer texts contain sufficient information to `predict' IQ scores and furthermore that IQ scores are a valid and relevant thing to measure about a person \citep{Bender:Medium:20}. Not only are these task formulations problematic, but as we describe in Section \ref{sec:du}, once sensitive data has been collected, it can be misused. 

\subsection{Collection, annotation, and documentation practices} 
\label{sec:doc}
A host of concerns regarding the practices of dataset collection, annotation, and documentation have been raised within recent years. In combination, these concerns reflect what  \cite{jo2020archives} describe as a \textit{laissez-faire} attitude regarding dataset development: rather than collecting and curating datasets with care and intentionality\,---\,as is more typical in other data-centric disciplines\,---\,machine learning practitioners have adopted an approach where anything goes. As one data scientist put it, ``if it is available to us, we ingest it'' \citep{Holstein2019}.

The common practices of scraping data from internet search engines, social media platforms, and other publicly available online sources faced significant backlash in recent years. For example, \cite{Prabhu2020LargeID} identified the presence of millions of non-consensual pornographic imagery within prominent computer vision datasets and facial analysis datasets have received push-back due to the inclusion of personal Flickr photos without data subject's knowledge \citep{difcritique}. In many instances, the legality of the data usage has come into question, as we discuss further in  Section \ref{sec:legal}. 

Dataset annotation practices have also come under great scrutiny within recent years. Much of this has focused on how subjective values, judgments, and biases of annotators contribute to undesirable or unintended dataset bias \citep{Ghai2020, Hube2019, van2016stereotyping, Misra2016, sap-etal-2019-risk}. More generally, several researchers have identified a widespread failure to recognize annotation work as \textit{interpretive work}, which in turn can result in a conflation of \textit{gold} labels in a collected dataset and \textit{real-world} objects, for which there may be no single ground truth label \citep{Miceli2020, Aroyo2015}.

Recent work by \citet{tsipras2020imagenet} has revealed that the annotation pipeline for ImageNet does not reflect the intention of its development for the purpose of object recognition in images. They note that ImageNet, constructed with the constraint of a single label per image, had its labels largely determined by crowdworkers indicating the visual presence of that object in the image. This has led to issues with how labels are applied, particularly to images with multiple objects, where the class of interest could include a background or obscured object that would be an unsuitable result for the image classification task of that particular photo. Furthermore, the nature of image retrieval for the annotation tasks biases the crowdworkers' response to the labeling prompt, making them much less effective at filtering out unsuitable examples for a class category. This is just one of several inconsistencies and biases in the data that hints at larger annotation patterns that mischaracterize the real world tasks these datasets are meant to abstractly represent, and the broader impact of data curation design choices in determining the quality of the final dataset.

Dataset documentation practices have also been a central focus, especially as dataset development processes are increasingly being recognized as a source of algorithmic unfairness. A recent study of publications that leverage Twitter data found data decisions were heavily under-specified and inconsistent across publications \citep{geiger2020}. \cite{Scheuerman2020} found a widespread under-specification of annotation processes relating to gender and racial categories within facial analysis datasets. Several dataset documentation frameworks have been proposed in recent years in an effort to address these concerns, with certain frameworks looking to not just capture characteristics of the output dataset but also report details of the procedure of dataset creation \citep{Gebru2018, bender2018data, Holland2018}.

The lack of rigorous and standardized dataset documentation practices has contributed to reproducibility concerns. For example, recent work undertook the laborious task of reconstructing ImageNet, following the original documented dataset construction process in an effort to test the generalization capabilities of ImageNet classifiers \citep{recht2019imagenet}. Despite mirroring the original collection and annotation methods\,---\,including leveraging images from the same time period\,---\,the newly constructed dataset was found to have different distributional properties. The differences were largely localized to variations in constructing ground truth labels from multiple annotations. More specifically, different thresholds for inter-annotator agreement were found to produce vastly different datasets, indicating yet again that ground truth labels in datasets do not correspond to truth.

This section centers on issues with criticisms of data themselves, and how representational issues, spurious correlations, problem legitimization, and the haphazard collection, annotation, and documentation practices are endemic to ML datasets. In the next section, we review ML methods which have been developed to deal with these issues.

\section{Filtering, augmenting, and other twists on datasets} 
\label{sec:use}

\begin{quote}
``[Sabotage is] the impossibly small difference between exceptional failures and business as usual, connected by the fact that the very same properties and tendencies enable either outcome.''

\,---\,Evan Calder Williams, `Manual Override'\footnote{\cite{williams2016sabotage}}
\end{quote}

Further insight into issues with dataset contents can be found in work that attempts to address these problems, still from within the same general paradigm. In this section, we survey recent work that proposes methods for exploring and adjusting datasets toward identifying and addressing some of the issues outlined in Section \ref{sec:design}. 

The massive sizes of contemporary machine learning datasets make it intractable to thoroughly scrutinize their contents, and thus it is hard to know where to begin looking for the kinds of representational and statistical biases outlined in the previous sections. While many of the biases discovered in datasets were found by using intuition and domain expertise to construct well-designed dataset ablations and audits, recent work has also proposed tools for using statistical properties of datasets to surface spurious cues and other issues with contents. The {\sc AfLite} algorithm proposed by \citet{Sakaguchi2020WINOGRANDEAA} provides a way to systematically identify dataset instances that are easily gamed by a model, but in ways that are not easily detected by humans. This algorithm is applied by \citet{lebras2020adversarial} to a variety of natural language processing datasets, and they find that training models on adversarially filtered data leads to better generalization to out-of-distribution data. Additionally, recent work proposes methods for performing exploratory data analyses based on training dynamics that reveal edge cases in the data, bringing to light labeling errors or ambiguous labels in datasets \citep{swayamdipta2020dataset}. These methods crucially rely on statistical patterns in the data to surface problem instances; it is up to human judgment to make sense of the nature of these problematic instances, whether they represent logical inconsistencies with the task at hand, cases of injustice, or both. As \citet{Denton2020BringingTP} propose in the `data genealogy' paradigm, we can qualitatively assess the design choices with respect to data sources, motivations, and methods used for constructing datasets.  \citet{Prabhu2020LargeID} and \citet{pipkin2020} show that meticulous manual audits of large datasets are compelling ways to discover the most surprising and disturbing contents therein; data-driven approaches may serve primarily as a map for where to begin looking. \citeauthor{pipkin2020} spent hundreds of hours watching the entirety of MIT's `Moments in Time' video dataset \citep{monfortmoments}. They provocatively point out through in their artistic intervention ``Lacework'' that the curators of massive datasets may have less intimate familiarity with the contents of these datasets than those who are paid to look at and label individual instances. 

In response to a proliferation of challenging perturbations derived from existing datasets to improve generalization capabilities and lessen the ability for models to learn shortcuts, \citet{liu-etal-2019-inoculation} propose `inoculation by fine-tuning' as a method for interpreting what model failures on perturbed inputs reveal about weaknesses of training data (or models). Recent papers also outline methodologies for leveraging human insight in the manual construction of counterfactual examples that complement instances in natural language processing datasets to promote better generalization \citep{gardner2020evaluating, kaushik2020contrast}.

The case of VQA-CP \citep{teney2020value} provides a cautionary tale of when a perturbed version of a dataset is, itself, prone to spurious cues. This complement to the original VQA dataset that consisted of instances redistributed across train and test sets was found to be easy to `solve' with randomly generated answers. Cleverly designed sabotages that are meant to strengthen models' ability to generalize may ultimately follow the same patterns as the original data, and are thus prone to the same kinds of artifacts. This has prompted attempts to make models more robust to any kind of dataset artifact, but also suggests that there is a broader view to be taken with respect to rethinking how we construct datasets for tasks overall. 

Considering that datasets will always be imperfect representations of real-world tasks, recent work proposes methods of mitigating the impacts of biases in data. \citet{teney2020learning} propose an auxiliary training objective using counterfactually labeled data to guide models toward better decision boundaries.  \citet{he-etal-2019-unlearn} propose the DRiFT algorithm for `unlearning' dataset bias. 
Sometimes, noise in datasets is not symptomatic of statistical anomalies or labeling errors, but rather, a reflection of variability in human judgment. \citet{pavlick2019inherent} find that human judgment on natural language inference tasks is variable, and that machine evaluation on this task should reflect this variability.

We emphasize that these procedural dataset modifications and bias mitigation techniques are only useful insofar as the dataset in question itself represents a valid task. In making lemonade from lemons, we must ensure the lemons are not ill-gotten or poorly formed.

\section{Dataset culture} 

\begin{quote}
``Data travel widely, but wherever they go, that’s where data are. For even when data escape their origins, they are always encountered within other significant local settings.''

-- Yanni Loukissas, \emph{Data Are Local}
\end{quote}

A final layer of critiques looks at the culture around dataset use in machine learning. In this section, we review papers that ask: What are issues with the broader culture of dataset use? How do our dataset usage, storage, and re-usage practices wrench data away from their contexts of creation? Lastly, what are the legal and privacy implications of these datasets?

\subsection{Benchmarking practices} 
\label{sec:bm}

Benchmark datasets play a critical role in orienting the goals of machine learning communities and tracking progress within the field \citep{Dotan2020, Denton2020BringingTP}.  Yet, the near singular focus on improving benchmark metrics has been critiqued from a variety of perspectives in recent years. Geoff Hinton has critiqued the current benchmarking culture, saying it has the potential to stunt the development of new ideas \citep{geoffbenchmarks}. Natural language processing researchers have exhibited growing concern with the singular focus on benchmark metrics, with several calls to include more comprehensive evaluations\,---\,including reports of energy consumption, model size, fairness metrics, and more\,---\,in additional to standard top-line metrics \citep{Ethayarajh2020, dodge2019, Schwartz2019}. \cite{Sculley2018WinnersCO} examine the incentive structures that encourage singular focus on benchmark metrics\,---\,often at the expense of empirical rigor\,---\,and offer a range of suggestions including incentivizing detailed empirical evaluations, including negative results, and sharing additional experimental details. From a fairness perspective, researchers have called for the inclusion of disaggregated evaluation metrics, in addition to standard top-line metrics, when reporting and documenting model performance \citep{modelcards}.  

The excitement surrounding leaderboards and challenges can also give rise to a misconstrual of what high performance on a benchmark actually entails. In response to the recent onslaught publications misrepresenting the capabilities of BERT language models, \cite{bender2020} encourage natural language processing researchers to be attentive to the limitations of tasks and include error analysis in addition to standard performance metrics. 

\subsection{Data management and distribution} 
\label{sec:dm}

Secure storage and appropriate dissemination of human-derived data is a key component of data ethics \citep{richards2014big}. To have a culture of care for the subjects of the datasets we make use of requires us to prioritize the well being of the subjects in the dataset throughout collection, development \emph{and} distribution. In order to do so systematically, the machine learning community still has much to learn from other disciplines with respect to how they handle the data of human subjects. Unlike in the social sciences or medicine, the machine learning field has yet to develop the data management practices required to store and transmit sensitive human data. 

\cite{metcalf2016human} go so far as to suggest the re-framing of data science as human subjects research, indicating the need for institutional review boards and informed content as researchers make decisions about other people's personal information. Particularly in consideration of an international context, where privacy concerns may be less regulated in certain regions,  the potential for the data exploitation is a real threat to the safety and well being of data subjects \citep{mohamed2020decolonial}. As a result, those that are the most vulnerable are at risk of losing control of the way in which their own personal information is handled. Without individual control of personal information, anyone who happens to be given the opportunity to access their unprotected data to can act with little oversight, potentially against the interests or wellbeing of data subjects. This can become especially problematic and dangerous in the most sensitive contexts of personal finance information, medical data or biometrics \citep{birhane2020algorithmic}. 

However, machine learning researchers developing such datasets rarely pay attention to this necessary consideration. Researchers will regularly distribute biometric information\,---\,for example, face image data\,---\,without so much as a distribution request form, or required privacy policy in place. Furthermore, the images are often collected without any level of informed consent or participation \citep{MegaPixels.cc, DIFnbc}. 

Even when these datasets are flagged for removal by the creators, researchers will still attempt to make use of that now illicit information through derivative versions and backchannels.  For example, \cite{pengface} finds that after certain problematic face datasets were removed, hundreds of researchers continued to cite and make use of copies of this dataset months later. Without any centralized structure of data governance for the research in the field, it becomes nearly impossible to take any kind of significant action to block or otherwise prevent the active dissemination of such harmful datasets.

\subsection{Use and Reuse}
\label{sec:du}

Several scholars have written on the importance of reusable data and code for reproducibility and replicability in machine learning \citep{stoddenBestPracticesComputational2014, stoddenDataScienceLife2020}. While the reuse of scientific data is often seen as an unmitigated good for scientific reproducibility \citep{pasquettoReuseScientificData2017}, here, we want to consider the potential pitfalls of taking data which had been collected for one purpose and using it for one in which it was not intended, particularly when this data reuse is morally and ethically objectionable to the original curators.  Science and technology scholars have considered the potential incompatibilities and reconstructions needed in using data from one domain in another \citep{edwardsVastMachineComputer2013}. Indeed, \citeauthor{strasserBigDataAnswer2017} discuss several major questions for big data in science and engineering, asking critically ``Who owns the data?'' and ``Who uses the data?'' \citeyearpar[p. 341-343]{strasserBigDataAnswer2017}. Although in Section \ref{sec:legal} we discuss ownership in a legal sense, ownership also suggests an inquiry into who the data have come from, such as the ``literal [\ldots]\ DNA sequences'' of individuals \citep[p. 342]{strasserBigDataAnswer2017} or other biometric information. In this case, considering data reuse becomes a pivotal issue of benchmark datasets.

Instances of data reuse in benchmarks are often seen in the scraping and mining context, especially when it comes to Flickr, Wikipedia, and other openly licensed data instances. Many of the instances in which machine learning datasets drawn from these and other sources which are serious privacy violations are well-documented by \cite{MegaPixels.cc}.

The reuse of data from one context to the context of  machine learning is exemplified well by historian of science Joanna Radin’s exploration of the peculiar history of the Pima Indians Diabetes Dataset (PIDD) and its introduction into the UCI Machine Learning Repository \citep{radinDigitalNativesHow2017}. The PIDD has been used thousands of times as a ``toy'' classification task and currently lives in the UCI repository, a major repository for machine learning datasets. The data were collected by the National Institutes of Health from the Indigenous community living at the Gila River Indian Community Reservation, which had been extensively studied and restudied for their high prevalence of diabetes. In her history of this dataset, Radin is attentive to the politics of the creation and processing of the data itself, rather than its deployment and use. The fact that ``data was used to refine algorithms that had nothing to do with diabetes or even to do with bodies, is exemplary of the history of Big Data writ large.'' \citeyearpar[p. 45]{radinDigitalNativesHow2017}. Moreover, the residents of the Reservation, who refer to themselves as the Akimel O'odham, had been the subject of intense anthropological and biomedical research, especially due to a high prevalence of diabetes, which in and of itself stemmed from a history of displacement and settler-colonialism. However, their participation in research had not yielded any significant decreases in obesity or diabetes amongst community members.

Another concerning example of data reuse occurs when derivative versions of an original dataset are distributed \,---\,beyond the control of its curators\,---\,without any actionable recourse for removal. The DukeMTMC (Duke Multi-Target, Multi-Camera) dataset was collected from surveillance video footage from eight cameras on the Duke campus in 2014, used without consent of the individuals in the images and distributed openly to researchers in the US, Europe, and China. After reporting in the \textit{Financial Times} \citep{murgiamadhumitaWhoUsingYour2019} and research by \citeauthor{MegaPixels.cc}, the dataset was taken down on June 2, 2019. However, \cite{pengFacialRecognitionDatasets2020} has recently highlighted how the dataset and its derivatives are still freely available for download and used in scientific publications. It is nearly impossible for researchers to maintain control of datasets once they are released openly or are not closely supervised by institutional data repositories.

\subsection{Legal issues}
\label{sec:legal}

A host of legal issues have been identified pertaining to the collection of benchmark datasets. Benchmarks are often mined from the internet, collecting data instances which have various levels of licensing attached and storing them into a single repository. Different legal issues arise at each stage in the data processing pipeline, from collection to annotation, from training to evaluation, from inference and the reuse of downstream representations such as word embeddings and convolutional features \citep{benjaminStandardizationDataLicenses2019}. Legal issues also arise which impact a host of different people in the process, including dataset curators, AI researchers, copyright holders, data subjects (those people whose likenesses, utterances, or representations are in the data), and consumers (those who are not in the data but are impacted by the inferences of the AI system). Different areas of law can protect (and also possibly harm) each of the different actors in turn \citep{khanHanna2020}. 

Benchmark datasets are drawn from a number of different sources, each with a different configuration of copyright holders and permissions for their use in training and evaluation in machine learning models. For instance, ImageNet was collected through several image search engines where licensing/copyright restrictions on data instances in those images are unknown \citep{russakovskyImageNetLargeScale2015}. The ImageNet project does not host the images on their website, and therefore sidestep the copyright question by claiming that they operate like a search engine \citep[ftn. 36]{levendowski2018copyright}. PASCAL VOC was collected via the Flickr API, meaning that the images were all held through the Creative Commons license \citep{everinghamPascalVisualObject2010}. Open licenses like Creative Commons allow for training of machine learning models under fair use doctrine 
\citep{merkelyUseFairUse2019}. Faces in the Wild and Labeled Faces in the Wild were collected through Yahoo News, and via an investigation of the captions on the images we can see that the major copyright holders of those images are news wire services, including the Associated Press and Reuters \citep{bergNamesFacesNews2004}. Other datasets are collected in  studio environment, where images were taken by dataset curators and therefore are copyright holders, which avoids potential copyright issues.

US copyright law is not well-suited to cover the range of uses of benchmark datasets, and there is not much case law establishing precedent in this area.  Legal scholars have defended the use of copyrighted material in data mining and training models by suggesting that this material's usage is protected by fair use, since it entails the non-expressive use of expressive materials \citep{sag2019new}. \citet{levendowski2018copyright} has argued that copyright is actually a useful tool for battling algorithmic bias by offering a larger pool of works from which machine learning practitioners can draw from. She argues that, given that pre-trained representations like \texttt{word2vec} and other word embeddings suffer from gender and racial bias \citep{caliskanSemanticsDerivedAutomatically2017,packer2018TextEmbeddings}, and other public domain datasets are older or obtained through means likely to result in amplified representation of stereotypes and other biases in the data (e.g. the Enron text dataset), that using copyright data can battle biased datasets and be used their use would fall under copyright's fair use exception.

Even in cases in which all data were collected legally from a copyright perspective\,---\,such as through open licenses like Creative Commons\,---\,many downstream questions remain, including issues about privacy, informed consent, and procedures of opt-out \citep{merkelyUseFairUse2019}. Copyright guarantees are not sufficient protections for safeguarding privacy rights of individuals, as seen in the collection of images for the Diversity in Faces and MegaFace datasets \citep{DIFnbc, murgiamadhumitaWhoUsingYour2019}. Potential privacy violations arise when datasets contain biometric information which can be used to identity individuals, including faces, fingerprints, gait, and voice amongst others. However, at least in the US, there is no national-level privacy law which deals with biometric privacy. A patchwork of laws exist in Illinois, California, and Virginia which have the potential to safeguard the privacy of data subjects and consumer. However, only the Illinois Biometric Privacy law requires corporate entities to provide notice to data subjects and obtain their written consent \citep{khanHanna2020}.

The machine learning and AI research communities have responded to the crisis by attempting to outline alternatives to licensing which make sense for research and benchmarking practices more broadly. The Montreal Data License\footnote{\url{https://montrealdatalicense.com/}} outlines different contingencies for a particular dataset, including whether the dataset will be used in commercial versus non-commercial settings, whether representations will be generated from the dataset, whether users can annotate the label or use subsets of it, and more \citep{benjaminStandardizationDataLicenses2019}. This is a step forward in clarifying the different ways in which the dataset can be used once it has been collected, and therefore is a clear boon for AI researchers who create their own data instances, such as photos developed in a studio or text or captions written by crowdworkers. However, this does not deal with the larger issue of the copyright status of data instances scraped from the web, nor the privacy implications of those data instances.

In this section, we've shed light on issues around benchmarking practices, dataset use and reuse, and the legal status of benchmark datasets. These issues are more about the peculiar practices of data in machine learning culture, rather than the technical challenges associated with benchmark datasets. In this way, we want to highlight how datasets work \textit{as} culture\,---\,that is, ``not [as] singular technical objects that enter into many different cultural interactions, but\ldots\ rather [as] unstable objects, culturally enacted by the practices people use to engage with them'' \citep{seaver2017}. Interrogating benchmark datasets from this view requires us to expand our frame from simply technical aspects of the system, to thinking how datasets intersect with communities of practice, communities of data subjects, and legal institutions \citep{selbst2019fairness}.

\section{Conclusion}

\begin{quote}
``Not all speed is movement.''

-- Toni Cade Bambara, `On the Issue of Roles'\footnote{\citet{bambara1970issue}}
% https://thefeministwire.com/2014/11/black-feminist-foremothers/
\end{quote}

In this paper, we have presented a survey of issues in dataset design and development,  as well as reflections on the broader culture of dataset use in machine learning. A viewpoint internal to this culture values rapid and massive progress: ever larger training datasets, used to train ever larger models, which post ever higher scores on ever harder benchmark tasks. What emerges from the papers we survey, however, is a viewpoint, largely external to that current culture of dataset use, which reveals intertwined scientific and ethical concerns appealing to a more careful and detail-oriented strategy.

Critiques of dataset design and development, especially of datasets that achieve the requisite size via opportunistic scraping of web-accessible (but not necessarily freely reusable) resources, point up various different kinds of pitfalls: First
there are pitfalls of representation wherein datasets are biased both in terms of which data subjects are predominantly included and whose gaze is represented. Second, we find pitfalls of artifacts in the data, which machine learning models can easily leverage to `game' the tasks. Third, we find evidence of whole tasks which are spurious, where success is only possible given artifacts because the tasks themselves don't correspond to reasonable real-world correlations or capabilities. Finally, we find critiques of insufficiently careful data annotation and documentation practice, which erode the foundations of any scientific inquiry based on these datasets. 

Attempts to rehabilitate datasets and/or models starting from the flawed datasets themselves further reinforce the problems outlined in the critiques of dataset design and development. The development of adversarial datasets or challenge sets, while possibly removing some spurious cues, doesn't address most of the other issues with either the original datasets or the research paradigm.

Critiques of the dataset culture itself focus on the overemphasis on benchmarking to the exclusion of other evaluation practices, data management and distribution, ethical issues of data reuse, and legal issues around data use. Hyper-focus on benchmarking pushes out work that connects models more carefully to their modeling domain and approaches not optimized for the available crop of benchmarks. The papers we surveyed suggest a need for work that takes a broader view than is afforded by the one-dimensional comparison of systems typical of benchmarks. Furthermore, critiques of data management and distribution show the need for growing a culture of care for the subjects of datasets in machine learning, i.e.\ to keep in mind that `data are people' and behave appropriately towards the people from whom we collect data \citep{raji2020discomfort}. Reflections of issues of data reuse emphasize the connection between data and its context, and the risks of harm (to data subjects and others) that arise when data is disconnected from its context and carried to and recontextualized in new domains. Finally, we surveyed papers exploring the  legal vulnerabilities inherent to current data collection and distribution practices in ML.

What paths forward are visible from this broader viewpoint? We argue that fixes that focus narrowly on improving datasets by making them more representative or more challenging might miss the more general point raised by these critiques, and we'll be trapped in a game of dataset whack-a-mole rather than making progress, so long as notions of `progress' are largely defined by performance on datasets.  At the same time, we wish to recognize and honor the liberatory potential of datasets, when carefully designed, to make visible patterns of injustice in the world such that they may be addressed (see, for example, the work of Data for Black Lives\footnote{\url{https://d4bl.org/}}). Recent work by \citet{register2020learning} illustrates how educational interventions that guide students through the process of collecting their own personal data and running it through machine learning pipelines can equip them with skills and technical literacy toward self-advocacy\,---\,a promising lesson for the next generation of machine learning practitioners and for those impacted by machine learning systems. 

In closing, we advocate for a turn in the culture towards carefully collected datasets, rooted in their original contexts, distributed only in ways that respect the intellectual property and privacy rights of data creators and data subjects, and constructed in conversation with the relevant scientific and scholarly fields required to create datasets that faithfully model tasks and tasks which target relevant and realistic capabilities. Such datasets will undoubtedly be more expensive to create, in time, money and effort, and therefore smaller than today's most celebrated benchmarks. This, in turn, will encourage work on approaches to machine learning (and to artificial intelligence beyond machine learning) that go beyond the current paradigm of techniques idolizing scale. Should this come to pass, we predict that machine learning as a field will be better positioned to understand how its technology impacts people and to design solutions that work with fidelity and equity in their deployment contexts.

\setcitestyle{numbers} 
\bibliographystyle{plainnat}
\bibliography{bibliography}

\end{document}